\documentclass[10pt,twocolumn,letterpaper]{article}

\usepackage{cvpr} 

\usepackage{pifont}
\usepackage{graphicx}
\usepackage{amsmath}
\usepackage{amssymb}
\usepackage{booktabs}
\usepackage{multirow}
\usepackage{colortbl}
\usepackage[dvipsnames]{xcolor}
\definecolor{tabtitle}{gray}{.8}
\definecolor{ours}{gray}{.95}
\definecolor{reda}{RGB}{208, 16, 76}
\definecolor{redb}{RGB}{217,148,143}
\definecolor{myyellow}{RGB}{190,144,0}
\definecolor{mygreen}{RGB}{30,144,255}
\definecolor{myblue}{RGB}{163,218,255}
\definecolor{LightCyan}{RGB}{224,255,255}

\newcommand{\none}{—}

\usepackage{algorithm}
\usepackage{algpseudocode}

\usepackage[pagebackref,breaklinks,colorlinks]{hyperref}
\newcommand{\mbf}{\mathbf}
\usepackage{bm}

\usepackage[capitalize]{cleveref}
\crefname{section}{Sec.}{Secs.}
\Crefname{section}{Section}{Sections}
\Crefname{table}{Table}{Tables}
\crefname{table}{Tab.}{Tabs.}

\begin{document}

\title{DQnet: Cross-Model Detail Querying for Camouflaged Object Detection}

\author{
Wei Sun$^1$
\quad
Chengao Liu$^1$
\quad
Linyan Zhang$^1$
\quad
Yu Li$^1$
\quad
Pengxu Wei$^3$
\quad
Chang Liu$^2$
\\
Jialing Zou$^1$
\quad
Jianbin Jiao$^1$
\quad
Qixiang Ye$^1$
\\
$^1$University of Chinese Academy of Sciences
\quad
$^2$Tsinghua University
\\
$^3$Sun Yat-Sen University
\\
}

\maketitle

\begin{abstract}Camouflaged objects are seamlessly blended in with their surroundings, which brings a challenging detection task in computer vision. Optimizing a convolutional neural network (CNN) for camouflaged object detection (COD) tends to activate local discriminative regions while ignoring complete object extent, causing the partial activation issue which inevitably leads to missing or redundant regions of objects. In this paper, we argue that partial activation is caused by the intrinsic characteristics of CNN, where the convolution operations produce local receptive fields and experience difficulty to capture long-range feature dependency among image regions. In order to obtain feature maps that could activate full object extent, keeping the segmental results from being overwhelmed by noisy features, a novel framework termed Cross-Model Detail Querying network (DQnet) is proposed. It reasons the relations between long-range-aware representations and multi-scale local details to make the enhanced representation fully highlight the object regions and eliminate noise on non-object regions. Specifically, a vanilla ViT pretrained with self-supervised learning (SSL) is employed to model long-range dependencies among image regions. A ResNet is employed to enable learning fine-grained spatial local details in multiple scales. Then, to effectively retrieve object-related details, a Relation-Based Querying (RBQ) module is proposed to explore window-based interactions between the global representations and the multi-scale local details. Extensive experiments are conducted on the widely used COD datasets and show that our DQnet outperforms the current state-of-the-arts. 

\end{abstract}

\begin{figure}[t]
\centering
\includegraphics[width=1.0\columnwidth]{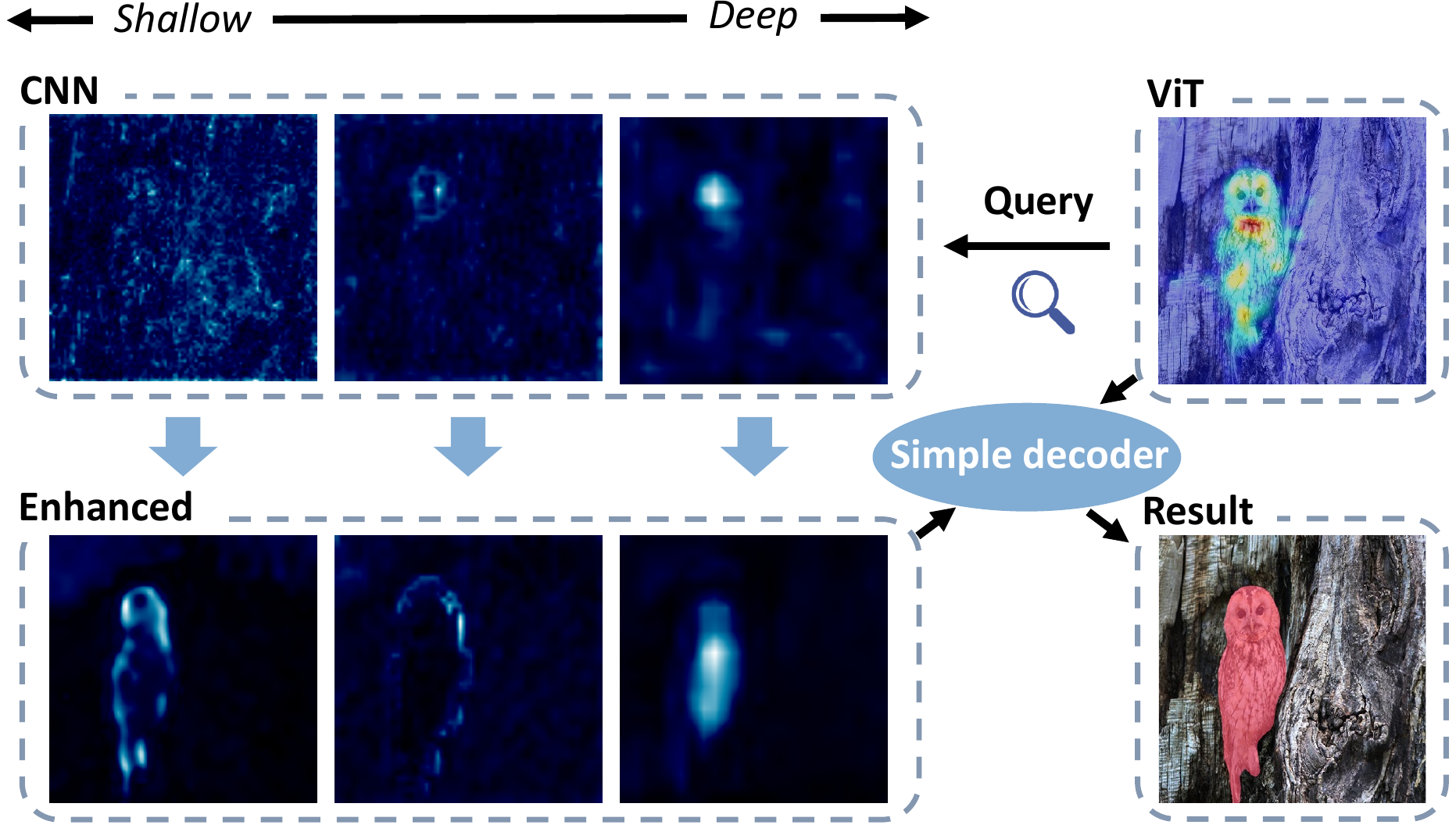}
\caption{Illustration of DQnet. With the contextual representation of ViT as global cues, our model queries crucial local details from the multi-scale CNN features. The enhanced representations have clear boundaries as well as few background noise, corresponding well with underlying camouflaged objects. (Best viewed in color.)}
\label{fig:first}
\end{figure}

\section{Introduction}
\label{sec:intro}

In nature, animals try to conceal themselves by adapting the color, shape, texture of their bodies to their surroundings, which helps them avoid being  hunted by predators. The same scheme is used in some artificial applications just as the wild animal photographing and mitilitary camouflage. Those camouflage strategies works by deceiving the visual perceptual system\cite{stevens2009animal}.  Recent algorithms designed for general object detection or salient object detection do not perform well in this difficult scenario.

Hence, camouflaged object detection (COD) is a challenging yet an important scientific topic. Solving the problem could benefit a lot of applications in computer vision, such as polyp segmentation\cite{fan2020pra_pranet}, lung infection segmentation\cite{fan2020inf}, photo-realistic blending, and recreational art.

The key issue of COD is that how to eliminate the ambiguities caused by the high intrinsic similarities between the target objects and the backgrounds, which demands learning representations precisely activating full objects extent. Recently, algorithms for COD based on deep CNNs have been explored a lot. However, due to the local receptive fields of CNN, it tends to activate local semantic elements of object or even some non-object regions, obtaining missing and redundant parts in COD results. Recently, much effort has been made to solve this problem for COD by proposing auxiliary tasks to augment the representations,  ({\em i.e.}, features for classification\cite{le2019ANet}, identification\cite{Fan_2020_CVPR_sinet}, or edge detection\cite{zhai2021mutual}. However, none of those work have paid attention to fundamentally solving the inherent defects of CNN’s local representation. Capturing the long-range feature dependency, which can be interpreted as the semantic relations between features in different spatial locations, is critical for the COD.

Recently,vision transformers~\cite{ViT_dosovitskiy2021an,liu2021swin,carion2020end,strudel2021segmenter} have shown great potential in many vision tasks. The design of image as patches facilitates the learning of object-centric representations, as evidenced by recent works, \emph{e.g.} DINO, EsViT, MAE~\cite{caron2021emerging,li2021efficient,he2022masked}. However, interestingly, we find that directly using ViT in COD yields unsatisfactory results. We argue that vision transformer cannot be directly mitigated to COD for the following two reasons: (1) In Multilayer Perceptron (MLP) layer, consecutive global matrix multiplications gradually wash out the low-level features, which is crucial to dense prediction tasks like COD. (2) Image patches are similar in COD scenario, and exclusively focusing on modeling the relationships among patches as in the ViT aggravates the already tense object-background ambiguity.

In this study, we present a simple yet effective framework termed Cross-Model Detail Querying network (DQnet), making the first attempt for COD with vanilla ViT. \textbf{Firstly}, a vanilla Vision Transformer(ViT) is introduced in DQnet, which is pretrained in a self-supervised manner\cite{he2022masked},  to generate representations with long-range correlations. 
\textbf{Secondly}, a plain CNN is adopted in DQnet to extract low-level visual cues within different scales which can well remedy fine-grained spatial details which have been gradually washed out during ViT inferences.
\textbf{Thirdly}, to retain crucial object-related details while discarding the noisy part, the CNN features are tokenized and incorporated with the ViT patches through a novel module named Relation-Based Querying (RBQ). It helps the DQnet progressively reason the relations between global-aware semantics and multi-scale details to distill the object-related visual cues for representation enhancement. It should be noted that the RBQ module is realized by window-based cross attention, which helps the relation reasoning focusing on spatial neighboring elements. It facilitates the feature enhancement with small memory heads, compared to full attention. As can be seen, the representation after enhancement in RBQ yields clusters of high purity within the target object region, and corresponds well with underlying object categories, \cref{fig:first}.

Our main contributions can be summarized as follows:
\begin{itemize}
	\item  We propose the first cross-model framework named DQnet for camouflaged object detection which uses cross-model representation to capture both continuous semantics and fine-grained details.
	
	\item  We introduce the novel Relation-Based Querying (RBQ) module to progressively reason the relations between global representations and the local details, which allows CNN to not only inherit global cues but also remedy the limited locality of ViT. Moreover, we design the Significance-Aware-Loss to further promote the multi-scale detail querying.
	
	\item  Our DQnet achieves SOTA performance on a variety of benchmarks, including CHAMELEON, CAMO, COD10K, and NC4K, outperforms existing COD models by a large margin.
\end{itemize}

\section{Related Work}
\noindent\textbf{Camouflaged Object Detection (COD)} 
Recently, the deep CNN networks with large capacity have been widely used to recognize the complex camouflaged objects.
Le \emph{et al.} \cite{le2019ANet} propose an end-to-end network called the ANet, adding a classification stream to a salient object detection model to boost the segmentation accuracy.
Yan \emph{et al.}\cite{yan2021mirrornet} find that horizontally flipped images could provide vital cues, and provide a two-stream MirrorNet with the original and flipped images as inputs.
Pei \emph{et al.} \cite{pei2022osformer} propose OSFormer, the first one-stage transformer framework for COD, and design a reverse edge attention module to focus on the edge features.
Sun \emph{et al.} use a multi-scale channel attention module to fuse the cross-level features in C$^2$FNet\cite{sun2021c2fNet}, and  employ object-related edge semantics to boost the performance of COD in BGNet\cite{sun2022boundary}.
Some bio-inspired frameworks \cite{Fan_2020_CVPR_sinet,mei2021camouflaged} are proposed by imitating the hunting process to gradually locate and search for the camouflaged object.
Pang \emph{et al.} \cite{pang2022zoomnet} get inspiration from human behavior. They zoom the original images to three scales and extract features separately to obtain more clues about the camouflaged objects.
Zhai \emph{et al.} \cite{zhai2021mutual} transform the camouflage feature maps and edge feature maps into sample-dependent semantic graphs to capture the high-level dependencies between them.
Some uncertainty-aware methods \cite{li2021ujsc, yang2021uncertainty} are also used to detect camouflaged objects.

\noindent\textbf{Transformers for vision.} 
More recently, vision transformers (ViTs) have become the new paradigm for visual recognition and have made great progress in a broad range of visual recognition tasks~\cite{ViT_dosovitskiy2021an,wang2021pyramid,liu2021swin}. Several properties of ViT make it a compelling model choice for visual representation learning. First, the self-attention mechanism in ViT offers a strong relational inductive bias, explicitly modeling the relations between input entities. Second, the design of image as patches facilitates the learning of object-centric representations, as evidenced by recent works, \emph{e.g.} DINO, EsViT, MAE~\cite{caron2021emerging,li2021efficient,he2022masked}, that demonstrate ViTs trained with self-supervised learning (SSL) capture objects in the image without label annotations.

\noindent\textbf{CNNs with non-local cues for segmentation.}
Various studies have attempted to integrate non-local mechanisms into CNNs by modeling global interactions of all pixels based on the feature maps. For instance, Wang \emph{et al.}~\cite{xwang18} designed a non-local operator, which could be plugged into multiple intermediate convolution. Built upon the encoder-decoder u-shaped architecture, Chen~\emph{et al.}~\cite{transunet} proposed to encode CNN feature maps into tokenized image patches, followed by a transformer to extract global context for promoting medical segmentation. Xi~\emph{et al.}~\cite{xie2022dual} used both Resnet and hierarchical Transformer as feature encoders, to extract     features with dual spatial sizes in parallel for high-resolution saliency detection.
Different from these approaches, we employ non-hierarchical vanilla ViT which provides single-stride non-local cues to effectively query details in an iterative way.

\noindent\textbf{Local Attention.}
Recently, a series of works re-introduce local attention to ViT for reducing computing complexity and facilitating hierarchical design for dense prediction tasks. Swin Transformer \cite{liu2021swin} computes self-attention in shifted windows, which implicitly built connections between non-overlapped windows, and has linear complexity with image size. Cswin\cite{dong2022cswin} designs a cross-shaped window attention, MSG-Transformer\cite{fang2022msg} uses additional tokens to exchange visual information across regions, and VSA\cite{zhang2022vsa} employs a window regression module to learn a varied-size Window. 
Based on local attention, we study the window mechanism and mix it with cross attention to facilitate our RBQ module design. Window attention helps the details querying focus on spatially neighboring patches, reducing the noise caused by object-background similarity.

\begin{figure*}[htbp]
  \centering
  \includegraphics[width=\linewidth]{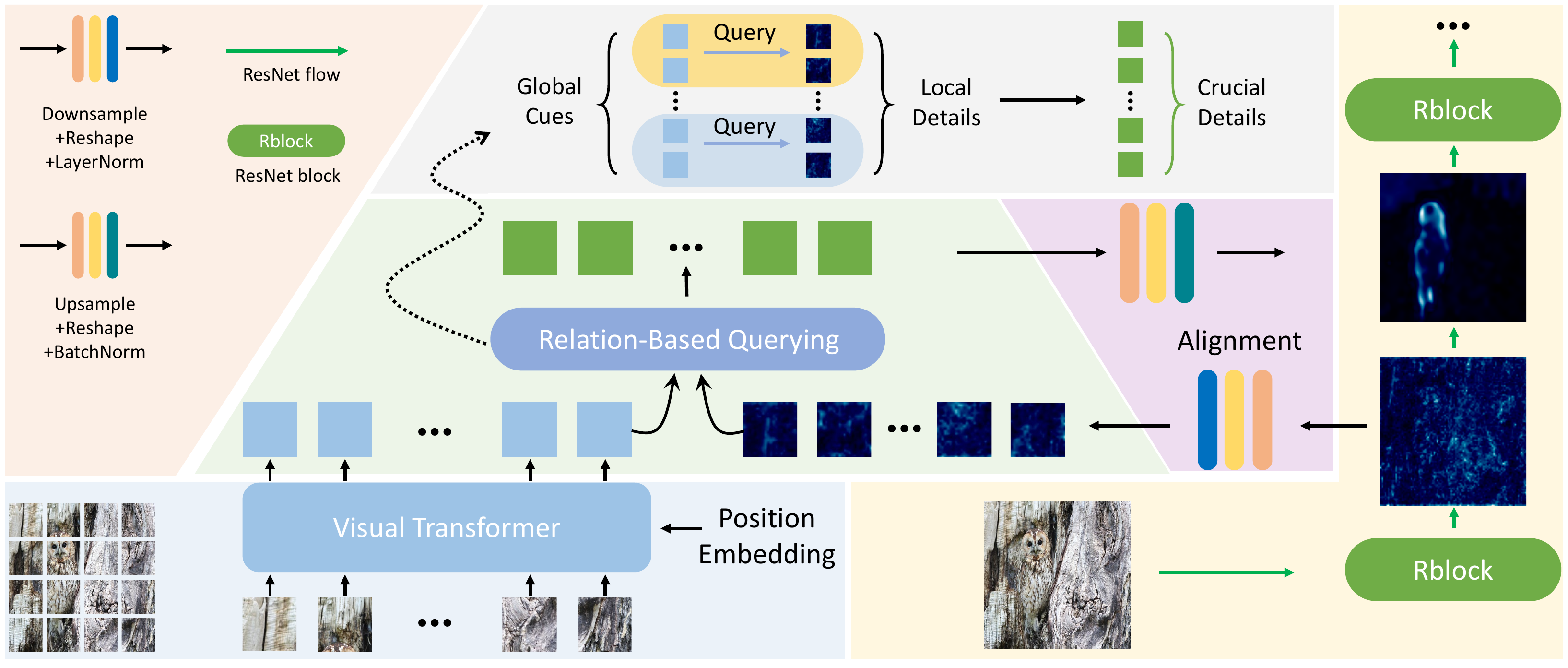}
  \caption{An overview of proposed DQnet. Vision transformer is used to provide global cues. ResNet is used to extract local details in different scales. Relation-Based Querying (RBQ) contains a windowed cross attention to query crucial fine-grained details and enhance multi-scale features. Feature alignment is applied before and after RBQ in each scale to eliminate the dimension inconsistency.}
  \label{fig:frame}
\end{figure*}
\label{sec:related}

\section{Methodology}
\subsection{Overall Architecture}
The architecture of proposed network is shown in \cref{fig:frame}. As can be seen, DQnet consists of two branches performing different tasks respectively: 1) CNN branch, which gradually encodes the input images to get a multilevel feature representation while keeping rich local details; 2) Transformer branch, where it extracts high-level semantic information in a cascaded self-attention manner. Moreover, we propose a Relation-Based Querying (RBQ) module to retrieve detailed local cues in multiple scales. Then, the enhanced multi-scale feature maps are combined to generate the segmentation using a light-weight decoder head. Below we provide the details of these proposed schemes.

\subsection{Feature Extractors}
\noindent\textbf{ViT Branch}
We will first briefly review the vision transformer structure in the original ViT\cite{ViT_dosovitskiy2021an}.  A raw image $\mbf{\hat{x}} \in \mathbb{R}^{H \times W \times C}$ is split into a sequence of patches $\mbf{x} = [x_1, ..., x_N] \in \mathbb{R}^{N \times P^2 \times C}$ where $(P, P)$ is the patch size, $N = HW/P^2$ is the number of patches and $C$ is the number of channels.  Each patch is flattened into a 1D vector and then linearly projected to a patch embedding to produce a sequence of patch embeddings $\mbf{x_0} = [\mbf{E}x_1, ..., \mbf{E}x_N] \in \mathbb{R}^{N \times D}$ where $\mbf{E} \in \mathbb{R}^{D \times (P^2C)}$. To capture positional information, learnable position embeddings $\mbf{pos} = [\text{pos}_1, ..., \text{pos}_N] \in \mathbb{R}^{N \times D}$ are added to the sequence of patches to get the resulting input sequence of tokens $\mbf{z_0} = \mbf{x_0}+\mbf{pos}$.

A transformer \cite{Transformer_NIPS2017_Vaswani} encoder composed of $L$ layers is applied to the sequence of tokens $\mbf{z}_0$ to generate a sequence of contextualized encodings $\mbf{z}_L \in \mathbb{R}^{N \times D}$. A transformer layer consists
of a multi-headed self-attention (MSA) block followed by a point-wise MLP block of two layers with layer norm (LN) applied before every block and residual connections added after every block:
\begin{eqnarray} \mbf{a_{i-1}} &=& \text{MSA}(\text{LN}(\mbf{z_{i-1}})) + \mbf{z_{i-1}},\\ \mbf{z_{i}} &=& \text{MLP}(\text{LN}(\mbf{a_{i-1}})) + \mbf{a_{i-1}},
\end{eqnarray} 
where $i \in \{1, ..., L\}$. The self-attention mechanism is composed of three point-wise linear layers mapping tokens to intermediate representations, queries $\mbf{Q} \in \mathbb{R}^{N \times d}$, keys $\mbf{K} \in \mathbb{R}^{N \times d}$ and values $\mbf{V} \in \mathbb{R}^{N \times d}$.  Self-attention is then computed as follows
\begin{equation}
  \text{MSA}(\mbf{Q}, \mbf{K}, \mbf{V}) = \text{softmax}\left(\frac{\mbf{QK}^T}{\sqrt{d}}\right)\mbf{V}.
\end{equation}

We employ a vanilla ViT-B\cite{ViT_dosovitskiy2021an} as our transformer branch, which obtains a contextualized encoding sequence containing rich global-aware information which is crucial to roughly identifying the potential target objects. 

\noindent\textbf{CNN Branch}
As shown in \cref{fig:frame}, the ResNet-50 is chosen as our CNN branch, which by convention adopts the feature pyramid structure, where the resolution of feature maps decreases with network depth while the channel number increases. Five feature maps are generated in different stages of ResNet-50, denoted as $\left\{ \bm{R}_i|i=1,2,3,4,5 \right\}$. 

Visual transformers~\cite{ViT_dosovitskiy2021an} project  an image patch into a vector through a single step, causing the lost of local details. In CNN, convolution kernels slide over feature maps with overlap, which preserves fine-grained local details yet lacks the ability to get continuous semantics. The transformer encoder on the other hand, is able to get accurate global-aware information. Combining these two together allows the feature extractor to not only inherit global information but also remedy the defects caused by ViT lacking spatial inductive bias. Consequently, the ViT branch is able to consecutively provide global cues for the ConNets branch.
\subsection{Cross-Model Querying}
Given the features from the CNN branch and the ViT branch, our goal is to combine them to get an enhanced representaitions and to enable explorations of effective feature enhancement for addressing the COD issues. To this end, we design a multi-scale detail querying mechanism to eliminate the inconsistency between these two kinds of features in an interactive way. 

\noindent\textbf{Feature Alignment.} 
To eliminate the dimension inconsistency, we first perform tokenization to the last four feature maps of ResNet (namely, $\left\{ \bm{R}_i|i=1,2,3,4 \right\}$) by reshaping the input $\bm{R}_i$ into a sequence of flattened 2D patches \{$\bm{Y}_i^k \in \mathbb{R}^{P^2 \cdot C}|k=1,..,N\}$, where each patch is of size $P \times P$ and $N=\frac{HW}{P^2}$ is the number of image patches (\emph{i.e.}, the input sequence length). Concretely, feature maps first require getting through maxpooling downsampling to align the spatial scales with the ViT patch embeddings. An 1$\times$1 convolution is then used to complete the channel dimension alignment, followed by a LayerNorm layer to regularize the features. Finally, the aligned features are sent to Relation-Based Querying (RBQ) module to get an enhanced representation for COD, which is illustrated in the next section. When fed back to the ResNet branch, the enhanced patch embeddings first get through deconvolution up-sampling to restore the spatial scale and to align the channel dimensions simultaneously, followed by a BatchNorm layer to recover the CNN-style regularization, as shown in \cref{fig:frame}.

It should be noted that the spatial and channel dimensions are constant for each ViT layer, whereas the classic CNN has a pyramid-like feature map structure where the spatial and channel dimensions vary in every stage. Due to this innate difference, determining which position to perform the alignment is an important issue. Here in our scenario, we simply use only the last feature map from the ViT to provide global cues, which should have the strongest features. The proposed alignment mechanism takes place at the beginning of the four ResNet stages, together with the RBQ to help retrieving crucial local details in every scale.

\begin{figure}[tbp]
  \centering
  \includegraphics[width=\linewidth]{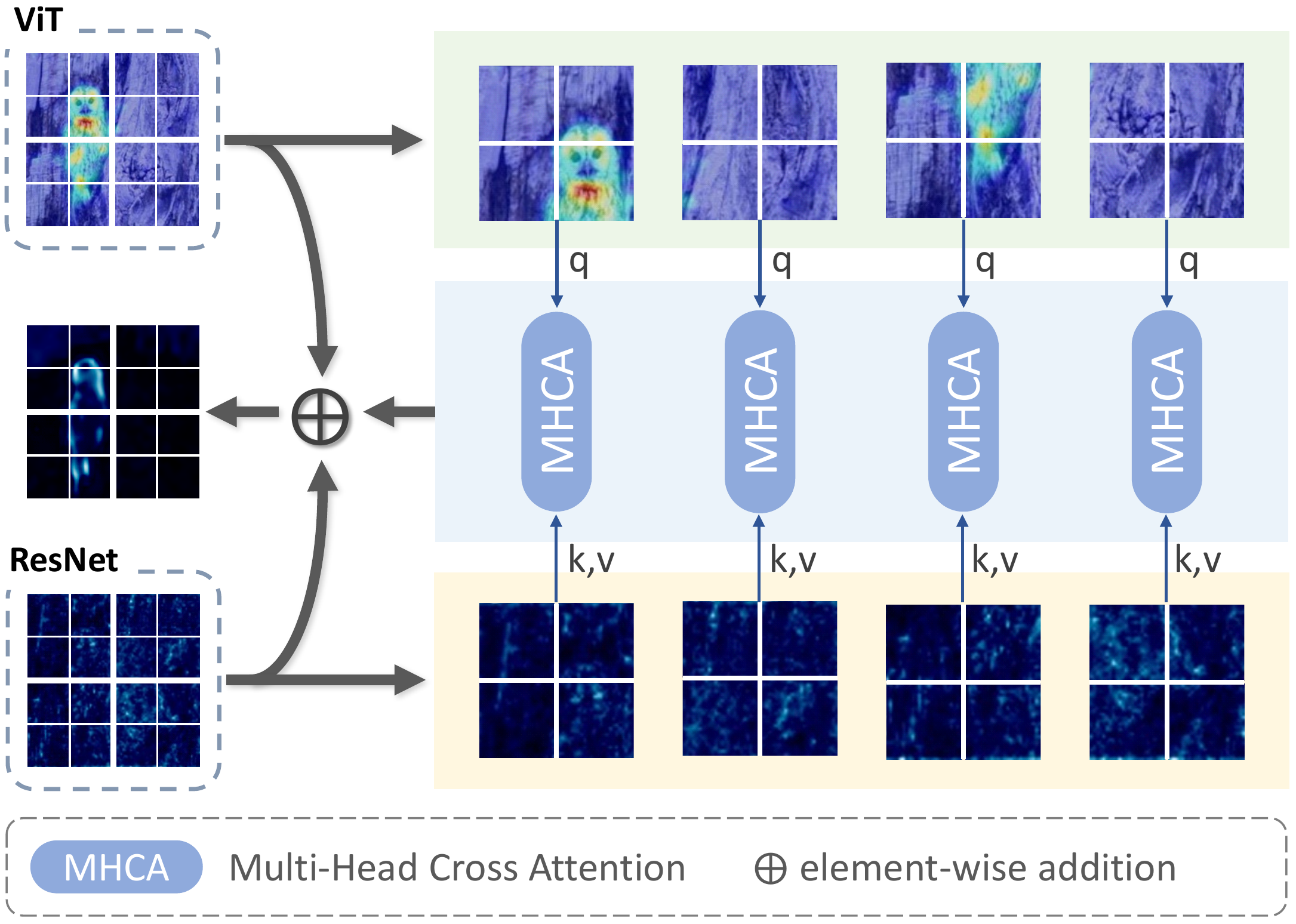}
  \caption{Illustration of the relation-based querying (RBQ). (Best viewed in color)}
 \label{fig:RBQ}
\end{figure}

\noindent\textbf{Relation-Based Querying.} 
 Conventionally, direct element-wise addition is used to fuse the aligned features. However, in COD scenario, directly fusion of low-level details and high-level semantics may lead to a phenomenon that the fine-graind details are easily overwhelmed by surrounding contextual information. Thus, we propose a Relation-Based Querying (RBQ) module to bridge the cross-model semantic gaps by calculating the pixel-wise relations of two kinds of features and adding the relation-based result to the original fusion. Given the ViT features $Z \in \mathbb{R}^{P^2 \cdot C}$ and tokenized ResNet features $Y \in \mathbb{R}^{P^2 \cdot C}$ as input, we employ window-based cross attention to reason the pixel-wise relations among spatial neighboring elements. Firstly, the features are partitioned into several non-overlapping windows, \ie, $\{Z_w^i \in \mathbb{R}^{w \times w \times C} | i \in [1, \dots, \frac{H\times W}{w^2}]\}$ and $\{Y_w^i \in \mathbb{R}^{w \times w \times C} | i \in [1, \dots, \frac{H\times W}{w^2}]\}$, where $w$ is the predefined window size. After that, the partitioned tokens are flattened along the spatial dimension and projected to query, key, and value tokens, \ie
\begin{equation} 
\begin{split}
\label{eq:qkv}
    Q_{w,f}^i &= Z_w^i\mathbf{W}_q^i, \\ K_{w,f}^i &= Y_w^i\mathbf{W}_k^i, \ \ \ V_{w,f}^i = Y_w^i\mathbf{W}_v^i, \\
\end{split}
\end{equation}
 where $\mathbf{W}_q$, $\mathbf{W}_k$, $\mathbf{W}_v$ $\in \mathbb{R}^{C\times C}$ are learnable parameters, $C$ is the embedding dimension, $\{Q_{w,f}^i, K_{w,f}^i, V_{w,f}^i \in \mathbb{R}^{w^2 \times N \times C'} | i \in [1, \dots, \frac{H\times W}{w^2}]\}$, where $Q,K,V$ represent the query, key, and value tokens, respectively, $N$ denotes the head number and $C'$ is the channel dimension along each head. Here the ViT feature is the only query as it provides powerful semantic guide for pixel-wise relation calculation. It is noted that $N \times C'$ equals the channel dimension $C$ of the given feature. Given the flattened query, key, and value tokens from the same default window, the window-based attention layers conduct full attention within the window, \ie,
\begin{equation}
    B_{w,f}^i = MHCA(Q_{w,f}^i, K_{w,f}^i, V_{w,f}^i).
\end{equation}
The $B_{w,f}^i \in \mathbb{R}^{w^2\times N \times C'}$ is the querying result after attention and $MHCA$ represents the vanilla multi-head cross-attention operation. The relative position embeddings are utilized during the attention calculation to encode spatial information into the features. The querying result $B_{w,f}^i$ are reshaped back to the window shape, \ie, $B_w^i \in \mathcal{R}^{w \times w \times C}$, and added with $Z_w^i$ and $Y_w^i$, \cref{fig:RBQ}. The same operation is individually repeated for each window and the generated features from all windows are then concatenated to recover the shape of input features, to get the final enhanced representation, $B \in \mathbb{R}^{P^2\times C}$. 

With the usage of window-based attention, the computational complexity decreases to linear to the input size, \ie, the computation complexity of RBQ module for each image is $\mathcal{O}(w^2HWC)$, compared to $\mathcal{O}((HW)^2C)$ of the global attention, which makes the detail querying more efficient.

It should be noted that we design RBQ to query details in different scales of CNN feature maps ($\left\{ \bm{s}_i|i=1,2,3,4 \right\}$), which progressively eliminates the imprecise feature entanglement when plainly adding two semantically-gapped features together. It seems that a full attention is also able to get the cross-model relationship. However, it aggravates the already tense object-background ambiguity and comes with large memory overhead when dealing with high resolution input. We claim that windowed attention is more suitable for detail querying purpose because the local windows help the relation reasoning focusing on spatial neighboring elements, which reduces the noise caused by object-background similarity. Our experiment in \cref{ablation} has proved that empirically.
\label{sec:RBQ}

\subsection{Loss Functions}
Following \cite{f3net}, we choose the weighted binary cross-entropy loss~($\mathcal{L}_\text{BCE}^w$) as our main loss. By calculating the difference between the center pixel and its surrounding environment, each pixel is assigned a different weight, so that the hard pixels can be obtained more attention. 

In fact, pixels located on difficult-to-detect regions (e.g., camouflage edge and occluded regions) are prone to activate differently in ViT feature and CNN features due to the much larger uncertainty. These pixels contain significant details which are critical to precise camouflage detection and deserve more attention. We argue that a larger weight value in these pixels could help the network focus on more siginificant features and better preserve the local details. Thus we propose the Significance Aware Loss (SAL) to supervise the detail querying results of RBQ, which can be defined as:

\begin{equation}
    \mathcal{L}_{SAL}=-\frac{\sum\limits^4_{s=1}\sum\limits^H_{i=1}\sum\limits^W_{j=1}(1+\beta\omega^s_{ij})\cdot \ell_{bce}(G^a_{ij},{\mathrm{P^s}}_{ij})}{\sum\limits^4_{s=1}\sum\limits^H_{i=1}\sum\limits^W_{j=1}(1+\beta\omega^s_{ij})} ,
    \label{eq:agl}
\end{equation}
where $\beta$ is a hyperparameter to adjust the impact of the weight $\omega^s_{ij}$ and $\ell_{bce}$ is the widely-used BCE loss, which is set to 2 in our experiment. $G_{ij}$ is the ground truth label of the pixel $(i,j)$, and $P^s_{ij}$ is the predicted probability by directly upsampling the enhanced features at the s-th layer of the ResNet. The weight $\omega^s_{ij}$ can be calculated by:

\begin{equation}\small
    \omega^s_{ij} = Sigmoid(Upsampling(\mid B^s\mid)),
    \label{eq:omega}
\end{equation}
where $B^s$ is the s-th enhanced representation as described in \cref{sec:RBQ}.

What's more, we also apply the weighted IoU loss to pay more attention to the global structure of the image as suggested by \cite{f3net}.
In the end, our total loss can be expressed as follow:
\begin{equation}
    \mathcal{L}_{total} = \mathcal{L}_\text{BCE}^w+\mathcal{L}_\text{IoU}^w+\mathcal{L}_{SAL},
    \label{eq:total}
\end{equation}
\label{sec:loss}

\begin{table*}[t]
	\centering
	\caption{
		Performance comparisons on common datasets.
		The best (second best) results are highlighted in {\color{reda} \textbf{red}} ({\color{mygreen} \textbf{blue}}).
		$^{\star}$: Using extra data.
	}
	\resizebox{\linewidth}{!}{%
		\begin{tabular}{l|cccc|cccc|cccc|cccc}
 \toprule[2pt]

                    \rowcolor{tabtitle}             & \multicolumn{4}{c|}{\textbf{CAMO}} & \multicolumn{4}{c|}{\textbf{CHAMELEON}} & \multicolumn{4}{c}{\textbf{COD10K}} & \multicolumn{4}{c}{\textbf{NC4K}}                                                                                                                                                                                                                                                                                                                                                                                                                                                                                                                                                                                   \\ \rowcolor{tabtitle}
 \multirow{-2}{*}{Model}         & S$_{m}$ $\uparrow$                 & F$^{\omega}_{\beta}$ $\uparrow$         & MAE $\downarrow$                                & E$_{m}$ $\uparrow$               & S$_{m}$ $\uparrow$               & F$^{\omega}_{\beta}$ $\uparrow$  & MAE $\downarrow$                            & E$_{m}$ $\uparrow$               & S$_{m}$ $\uparrow$               & F$^{\omega}_{\beta}$ $\uparrow$  & MAE $\downarrow$                           & E$_{m}$ $\uparrow$               & S$_{m}$ $\uparrow$               & F$^{\omega}_{\beta}$ $\uparrow$  & MAE $\downarrow$                            & E$_{m}$ $\uparrow$              \\ \midrule[1pt]

 ANet\-SRM~\cite{le2019ANet}           & 0.682                              & 0.484                                   & 0.126                                                            & 0.722                            & \none                            & \none                            & \none                            & \none                            & \none                            & \none                            & \none                            & \none                            & \none                            & \none                            & \none                            & \none                                                       \\
 SINet~\cite{Fan_2020_CVPR_sinet}             & 0.745                              & 0.644                                   & 0.092                                                            & 0.829                            & 0.872                            & 0.806                            & 0.034                                                        & 0.946                            & 0.776                            & 0.631                            & 0.043                                            & 0.874                            & 0.808                            & 0.723                            & 0.058                                        & 0.883                            \\
 SLSR~\cite{lv2021simultaneously}                & 0.787                              & 0.696                                   & 0.080                                                           & 0.854                            & 0.890                            & 0.822                            & 0.030                             & 0.948  & 0.804                            & 0.673                            & 0.037                                                        & 0.892  & 0.840  & 0.766 & 0.048    &  0.907 \\
 MGL-R~\cite{zhai2021mutual}            & 0.775                              & 0.673                                   & 0.088                                                            & 0.842                            & 0.893 & 0.812                            & 0.031                     & 0.941                            & 0.814  & 0.666                            & 0.035              & 0.890                            & 0.833                            & 0.739                            & 0.053                                            & 0.893                            \\
 PFNet~\cite{mei2021camouflaged}          & 0.782                              & 0.695                                   & 0.085                                                            & 0.855                            & 0.882                            & 0.810                            & 0.033                                          & 0.945                            & 0.800                            & 0.660                            & 0.040                                    & 0.890                            & 0.829                            & 0.745                            & 0.053                                       & 0.898                            \\
 UJSC$^{\star}$~\cite{li2021ujsc}      & 0.800   &  0.728        &  0.073      &  0.873 & 0.891  & 0.833 & 0.030   &  0.955 & 0.809                            &  0.684  &  0.035    & 0.891                            & 0.842 &0.771 &  0.047  &  0.907 \\
 MirrorNet~\cite{yan2021mirrornet}  & 0.785                              & 0.719         &  0.077                                 & 0.850                            & \none                            & \none                            & \none                            & \none                            & \none                            & \none                            & \none                            & \none                                            & \none                            & \none                            & \none                            & \none                            \\
 C$^2$FNet~\cite{sun2021c2fNet}        &  0.796    &  0.719         & 0.080                               &  0.864  & 0.888                            &  0.828  & 0.032                               & 0.946                            & 0.813                            &  0.686 &  0.036   &  0.900 & 0.838                            & 0.762                            & 0.049                                                &  0.904  \\
 UGTR~\cite{yang2021uncertainty}            & 0.784                              & 0.684                                   & 0.086                                                            & 0.851                            & 0.888                            & 0.794                            &  0.031            & 0.940                            &  0.817 & 0.666                            &  0.036             & 0.890                            & 0.839                            & 0.746                            & 0.052                                   & 0.899                            \\
  BGNet~\cite{sun2022boundary}          &0.812  &0.749 &0.073 &0.870 &0.901 &{\color{mygreen} \textbf{0.850}} &0.027 &0.938 &0.831  &0.722 &0.033 &0.901 &0.851  &{\color{mygreen} \textbf{0.788}} &0.044 &0.907
 \\  
 ZoomNet~\cite{pang2022zoomnet}          & {\color{mygreen} \textbf{0.820}}      & {\color{mygreen} \textbf{0.752}}          & {\color{mygreen} \textbf{0.066}}            & {\color{mygreen} \textbf{0.892}}    &{\color{mygreen} \textbf{0.902}}    &  0.845    & {\color{mygreen} \textbf{0.023}}      &{\color{mygreen} \textbf{0.958}}    &  {\color{mygreen} \textbf{0.838}}    & {\color{mygreen} \textbf{0.729}}    &  {\color{mygreen} \textbf{0.029}}      &  {\color{mygreen} \textbf{0.911}}   & {\color{mygreen} \textbf{0.853}}    &  0.784    &  {\color{mygreen} \textbf{0.043}}        &  {\color{mygreen} \textbf{0.912}} \\  \rowcolor{LightCyan} 
  DQnet(ours)                            &  {\color{reda} \textbf{0.898}}      & {\color{reda} \textbf{0.898}}           & {\color{reda}\textbf{0.034} }         & {\color{reda}\textbf{0.944}}    & {\color{reda}\textbf{0.915}}    & {\color{reda}\textbf{0.866}}    & {\color{reda}\textbf{0.021}}         & {\color{reda}\textbf{0.960}}    & {\color{reda}\textbf{0.882}}    & {\color{reda}\textbf{0.801}}    & {\color{reda}\textbf{0.021}}       & {\color{reda}\textbf{0.930}}    &{\color{reda}\textbf{0.901}}    & {\color{reda}\textbf{0.859}}    & {\color{reda}\textbf{0.029}}        & {\color{reda}\textbf{0.942}}    \\ \bottomrule[2pt]
\end{tabular}%

	}
	\label{tab:sota}
\end{table*}

\vspace{-2em}
\section{Experiments}
\subsection{Experimental Setup}
\noindent\textbf{Datasets.} We evaluate our method on four benchmark datasets: CHAMELEON \cite{skurowski2018animal_chameleon}, CAMO \cite{le2019ANet}, COD10K \cite{Fan_2020_CVPR_sinet} and NC4K~\cite{lv2021simultaneously}. CHAMELEON \cite{skurowski2018animal_chameleon} has 76 images collected from the Internet via the Google search engine using ``camouflaged animal'' as a keyword with hand-annotated labels. CAMO \cite{le2019ANet} contains 1,250 camouflaged images covering 8 categories(1,000 for training, 250 for testing). COD10K \cite{Fan_2020_CVPR_sinet} is a large-scale dataset including 5,066 camouflaged, 3,000 background and 1,934 non-camouflaged images, which provides high-quality fine annotation, reaching the level of matting. NC4K~\cite{lv2021simultaneously} is currently the largest benchmark testing dataset including 4,121 images from the Internet. We follow previous work ~\cite{lv2021simultaneously,pang2022zoomnet} to use the training set of CAMO \cite{le2019ANet} and COD10K \cite{Fan_2020_CVPR_sinet} as the training sets (4,040 images) and others as testing sets.

\noindent \textbf{Evaluation Metrics.}
To comprehensively compare our proposed model with other state-of-the-art methods, We use four widely used metrics to evaluate our method: structure-measure (S$_{m}$) \cite{fan2017structure_smeasure}, E-measure (E$_{m}$) \cite{fan2018enhanced_emeasure}, weighted F-measure (F$^{\omega}_{\beta}$) \cite{margolin2014evaluate_wfmeasure}, and mean absolute error (MAE).
Structure-measure (S$_{m}$) \cite{fan2017structure_smeasure}focuses on evaluating the structural information of the prediction maps, by computing the object-aware and region-aware structure similarities between the prediction and the ground truth. E-measure (E$_{m}$) \cite{fan2018enhanced_emeasure}is based on the human visual perception mechanism to assess the overall and local accuracy of COD results. F-measure (F$^{\omega}_{\beta}$) \cite{margolin2014evaluate_wfmeasure}is a comprehensive measure of both the precision and recall of the prediction map. The mean absolute error (MAE) metric is also used in our evaluation to measure the foreground-background segmentation error.

\noindent \textbf{Implementation Details.} Our method is implemented based on PyTorch. Two NVIDIA GeForce RTX 3090 GPU (with 24GB memory) is used for both training and testing.  For data preparation, we perform data augmentation techniques on all training data, including random flipping, random rotating and scaling in the range of [0.75, 1.25]. The parameters of ResNet-50 and ViT-B are both initialized with weights pre-trained on ImageNet while the remaining layers are initialized randomly. For effective finetuning, we set the layer decay to 0.75 for ViT. For optimization, we use the AdamW optimizer for loss optimization. We set the batch size to 12 and adjust the learning rate by the poly strategy \cite{liu2015parsenet} with the initial learning rate of 0.001 and the power of 0.9. Both the resizing processes use bicubic interpolation.

\begin{figure*}[htbp]
  \centering
  \includegraphics[width=\linewidth]{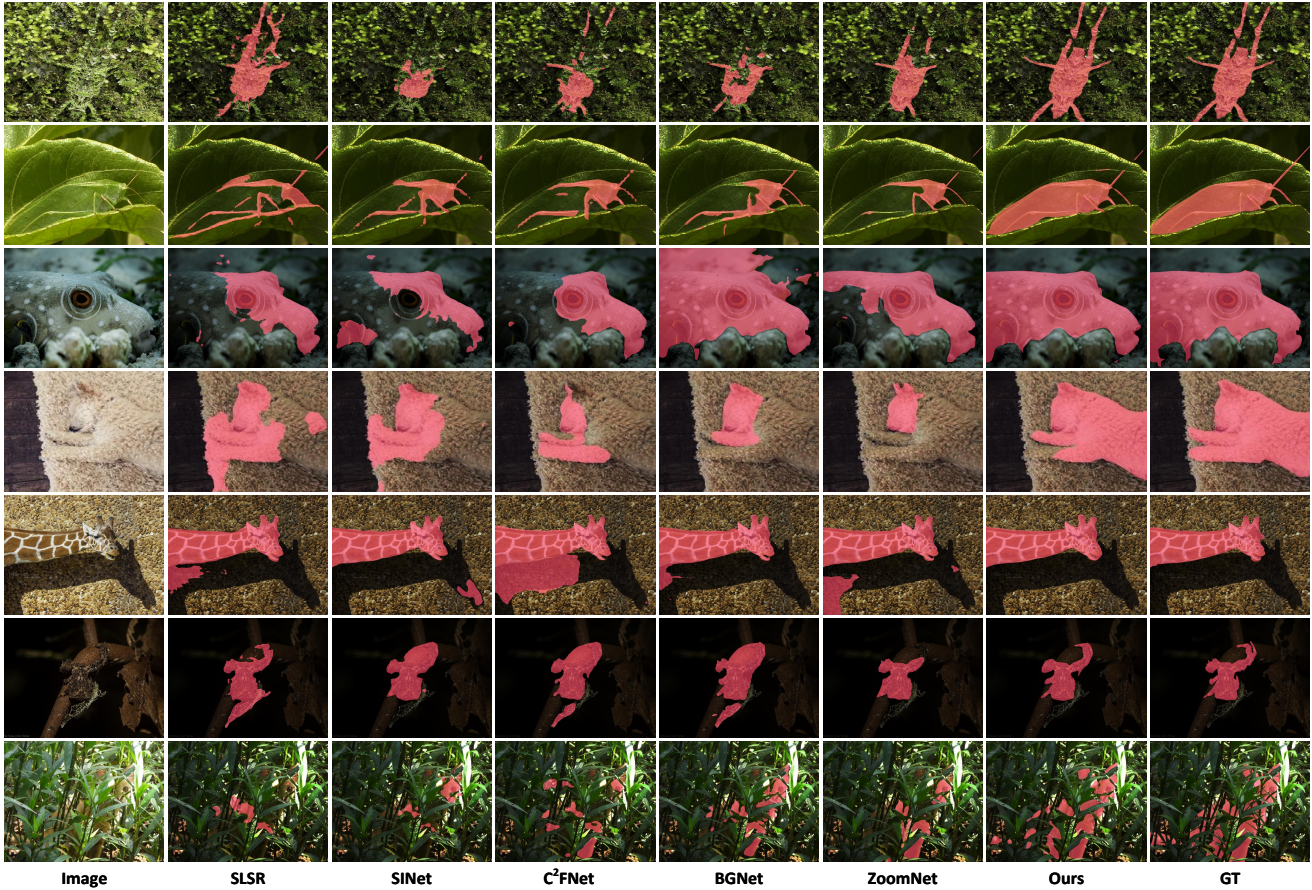}
  \caption{Visual comparisons of some recent COD methods and ours. (Best viewed in color with zoom in)}
 \label{fig:fig4}
\end{figure*}

\subsection{Comparision with State-of-the-art methods}
We compare our proposed FCNet with 11 SOTA methods, including ANet-SRM\cite{le2019ANet}, SINet\cite{Fan_2020_CVPR_sinet}, SLSR\cite{lv2021simultaneously}, MGL-R\cite{zhai2021mutual}, PFNet\cite{mei2021camouflaged}, UJSC\cite{li2021ujsc}, MirroNet\cite{yan2021mirrornet},
C$^2$FNet~\cite{sun2021c2fNet},
UGTR\cite{yang2021uncertainty},
BGNet\cite{sun2022boundary},
ZoomNet\cite{pang2022zoomnet} on four benchmark datasets.

\noindent\textbf{Quantitative Comparison.}
As can be seen in \cref{tab:sota}, even without using any extra dataset or auxiliary task, our method outperforms all the other methods by a large margin on all four standard metrics. Specifically, compared with the currently second best result, F$^{\omega}_{\beta}$ increased by $7.6\%$ on average, S$_{m}$ increased by $4.6\%$, E$_{m}$ increased by $2.6\%$, and MAE lowered by $1.4\%$. Notably, even on the currently most difficult COD dataset NC4K, our DQnet still achieves a $4.8\%$ gain in terms of S$_{m}$ and a remarkable $7.5\%$ gain in terms of F$^{\omega}_{\beta}$. Such a performance gain proves the importance of effective detail querying. Progressively retrieving crucial local details under strong global cues could be vital for full object extent activation and precise boundary delineation in the difficult COD scenario. 

\noindent\textbf{Visual Comparison.}
Visual comparisons of different COD methods on several typical samples are shown in \cref{fig:fig4}. While the SOTAs are typically confused by the background which shares similar context with camouflaged objects (\emph{e.g.}, first four rows) or resembles the target object due to shadow or illumination (\emph{e.g.}, 5-\emph{th} and 6-\emph{th} rows), our method can successfully infer the camouflaged region with complete object extent. This is mainly contributed by the proposed detail querying strategy which could help suppress the background noise and enhance the details within the target object region. Furthermore, benefited by the cross-model interaction in the  detail querying process, our method can capture fine-grained object details without losing the general shape of a specific category and thus has the ability to finely segment the camouflaged objects under severe occlusion. (\emph{e.g.}, 7-\emph{th} row)

\subsection{Model Evaluation and Analyses}\label{ablation}
In this section, we conduct extensive ablation analyses on different components. Because COD10K\cite{Fan_2020_CVPR_sinet} is the most widely-used large-scale COD dataset, and contains various objects and scenes, all subsequent ablation experiments are carried out on it.

\begin{figure*}[htbp]
  \centering
  \includegraphics[width=\linewidth]{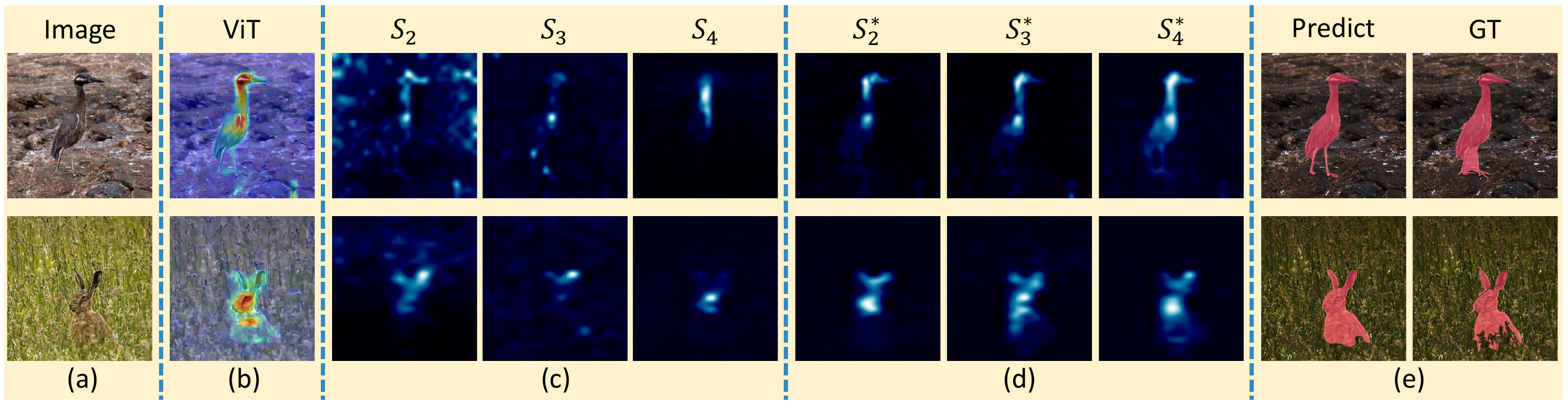}
  \caption{Visualization of mutil-scale detail querying. (a) Input image. (b) ViT attention map. (c) Vanilla ResNet features in the last three stages. (d) Enhanced ResNet features after RBQ. (e) Segmentation result and ground truth. (Best viewed in color)}
 \label{fig:feature}
\end{figure*}

\noindent\textbf{The Effectiveness of feature alignment}
To verify the effectiveness of proposed feature alignment, we use plain element-wise addition to perform feature enhancement after the feature alignment. When validating the vanilla ViT-B, we directly upsample the single-stride representation with deconvolution to get the predictions due to the non-hierarchical architecture of the vanilla ViT. For the vanilla ResNet-50 case, we use the latter four stage features to construct the feature pyramid and follow similiar simple decoder design with our method. We also construct the feature pyramid using both ViT-B and ResNet-50 to evaluate the model ensemble performance. The results are shown in \cref{tab:modules}. From \ding{172}-\ding{175}, it can be been that a simple ensemble only gives a minor gain compared with vanilla ViT-B, while feature alignment with a plain addition could make a significant contribution to the performance. This shows that an interactive combining after the feature alignment could inherit the strengths of both models more easily.

\noindent\textbf{The Effectiveness of RBQ and SAL}
As shown in \cref{tab:modules} \ding{176}, with RBQ, the value of F$^{\omega}_{\beta}$ increaced by 8.9\% compared to ViT-B and 17.8\% compared to Res-50. It indicates that high-level global semantics and low-level local details in multiple scales are both critical to accurate COD. Compared with vanilla ResNet features, RBQ effectively suppresses the noise caused by object-background similarity, obtaining features fully highlight the object regions, \cref{fig:feature}. Furthermore, as shown in \cref{tab:fusion}, RBQ effectively refines the feature enhancement by introducing multi-scale querying to fill the cross-model semantic gap progressively, which further improves the performance compared with element-wise operation and feature concatenation. Moreover, \cref{tab:modules} \ding{177} validates the effectiveness of the SAL design.

\begin{table}[tbp]
\caption{Comparison of different architectures and compositions}

\renewcommand\arraystretch{1}
\label{table:ablation1}
\centering
\resizebox{\linewidth}{!}{%
\begin{tabular}{l|cccc}
\hline
\multirow{2}{*}{Composition} & \multicolumn{4}{c}{COD10K} \\ \cline{2-5} 
\multicolumn{1}{c|}{}                           & S$_{m}$ $\uparrow$ &  F$^{\omega}_{\beta}$ $\uparrow$ & MAE $\downarrow$  & E$_{m}$ $\uparrow$  \\ \toprule
\ding{172}vanilla Res-50                            & 0.756  & 0.621  & 0.046  & 0.856 \\
\ding{173}vanilla ViT-B                        & 0.833  & 0.710  & 0.033  & 0.876 \\
\ding{174}emsemble           & 0.838  & 0.725  & 0.030  & 0.884  \\
\ding{175}feature alignment + element-wise addition   & 0.856  & 0.761  & 0.025  & 0.909  \\
\ding{176}feature alignment + RBQ      &0.874  &0.799  &0.022  &0.924 \\\rowcolor{LightCyan}
\ding{177}feature alignment + RBQ + SAL    &0.882 &0.801 &0.021 &0.930 \\ \hline
\end{tabular}%
}
\label{tab:modules}
\end{table}

\begin{table}[]
\caption{Ablation study on RBQ}
\renewcommand\arraystretch{1}
\label{table:ablation1}
\centering
\resizebox{\linewidth}{!}{%
\begin{tabular}{c|cccc}
\hline
\multirow{2}{*}{Window Size} &
\multicolumn{4}{c}{COD10K} \\  \cline{2-5}
                   & S$_{m}$ $\uparrow$ &  F$^{\omega}_{\beta}$ $\uparrow$ & MAE $\downarrow$  & E$_{m}$ $\uparrow$  \\ \toprule

element-wise addition             & 0.856  & 0.761  & 0.025  & 0.909  \\ 
element-wise multiply             & 0.858  & 0.758  & 0.025  & 0.913 \\ 
feature concatenation            & 0.847  & 0.755  & 0.026  & 0.904  \\ 
\rowcolor{LightCyan}
window-based attention (RBQ)& 0.882  & 0.801  & 0.021  & 0.930 
\\\hline

\end{tabular}%
}
\label{tab:fusion}
\end{table}

\noindent\textbf{The positioin of RBQ}
To investigate the impact of RBQ position on the COD performance, we conduct a series of experiments implementing RBQ in different stages of ResNet. As shown in \cref{tab:position}, starting with only enhancing the last stage of ResNet, the performance gradually improves as the number of enhanced stage increase until reaching the best, where all ResNet stages are enhanced with RBQ. This demonstrates the excellent scalability of the proposed RBQ module, which is capable of querying useful local details in different scales.

\noindent\textbf{Window based attention in RBQ}
In \cref{tab:window}, we show the effects of different window sizes in the proposed RBQ.
It can be seen from the results that the best performance appears when the window size is equal to 4. Also, it achieves a good balance between performance and efficiency. When replacing the window-based attention with full attention, the performance would decline to some extent. This is because that indiscriminately modelling pixel-wise relations aggravates the already tense object-background ambiguity and thus hinders querying fine-grained details. This validates the rationality of our design to perform details querying by focusing on spatially neighboring elements.

\begin{table}[tbp]
\caption{Performance with different position of RBQ. }
\label{table:V4}
\centering
\renewcommand\arraystretch{1}
\resizebox{\linewidth}{!}{%
\begin{tabular}{c|c|c|c|cccc}
\hline
\multicolumn{1}{c|}{\multirow{2}{*}{Stage1}} & \multicolumn{1}{c|}{\multirow{2}{*}{Stage2}} &\multicolumn{1}{c|}{\multirow{2}{*}{Stage3}} &\multicolumn{1}{c|}{\multirow{2}{*}{Stage4}} & \multicolumn{4}{c}{COD-10K} \\ \cline{5-8} 
                        &&&& S$_{m}$ $\uparrow$ &  F$^{\omega}_{\beta}$ $\uparrow$ & MAE $\downarrow$  & E$_{m}$ $\uparrow$       \\ \toprule
\none&\none&\none &   \checkmark    & 0.868     & 0.784     & 0.023    & 0.920            \\
\none  &  \none&\checkmark&\checkmark& 0.874     & 0.791 & 0.022    & 0.924            \\
 \none&\checkmark&\checkmark&\checkmark  & 0.880     & 0.797     & 0.021    & 0.926       \\\rowcolor{LightCyan}
\checkmark&\checkmark&\checkmark&\checkmark& 0.882  & 0.801  & 0.021  & 0.930           \\
 \hline
\end{tabular}%
}
\label{tab:position}
\end{table}

\begin{table}[]
\caption{Ablation study on window size}

\renewcommand\arraystretch{1}
\label{table:ablation1}
\centering
\resizebox{\linewidth}{!}{%
\begin{tabular}{c|c|cccc}
\hline
\multirow{2}{*}{Window Size} &
\multirow{2}{*}{Memory (M)} &
\multicolumn{4}{c}{COD10K} \\  \cline{3-6}
&                    & S$_{m}$ $\uparrow$ &  F$^{\omega}_{\beta}$ $\uparrow$ & MAE $\downarrow$  & E$_{m}$ $\uparrow$  \\ \toprule
1&0.48& 0.880  & 0.799  & 0.022  & 0.925\\
2 &2.00                        & 0.880  & 0.799  & 0.022  & 0.925 \\\rowcolor{LightCyan}
4 &9.52                    & 0.882  & 0.801  & 0.021  & 0.930 \\
6 &27.01                   & 0.880  & 0.798  & 0.022  & 0.926 \\
8 &62.43                   & 0.876  & 0.793 & 0.023  & 0.922\\
10 &127.08         & 0.872  & 0.789 & 0.023  & 0.920\\
\hline
full  &195.04& 0.871  & 0.788 & 0.023  & 0.917
\\\hline

\end{tabular}%
}
\label{tab:window}
\end{table}

\section{Conclusion}

In this paper, we propose the cross-model detail querying network (DQnet) for COD. DQnet takes full advantage of the cascaded self-attention mechanism in the visual transformer for long-range feature dependency extraction and object extent localization, as well as the local convolution mechanism in CNN for fine-grained details extraction. 
We propose the relation-based querying (RBQ) module to distill fine-grained details crucial for COD while disgrading the noisy part. Experiments on four widely-used benchmarks show that DQnet significantly improves the COD performance. As the first and solid COD baseline with cross-model framework, DQnet provides a fresh insight to the challenging COD problem by re-focusing on the feature extractor design. This makes our method compatible with COD decoder developments along various directions that are not necessarily limited by effective feature fusion, feature refinement, and auxiliary tasks design.

\section{Model Details}\label{sec:detailsaboutmodels}
\subsection{Decoder Framework}\label{sec:decoderframework}
\begin{figure}[htbp]
\centering
\includegraphics[width=1.0\columnwidth]{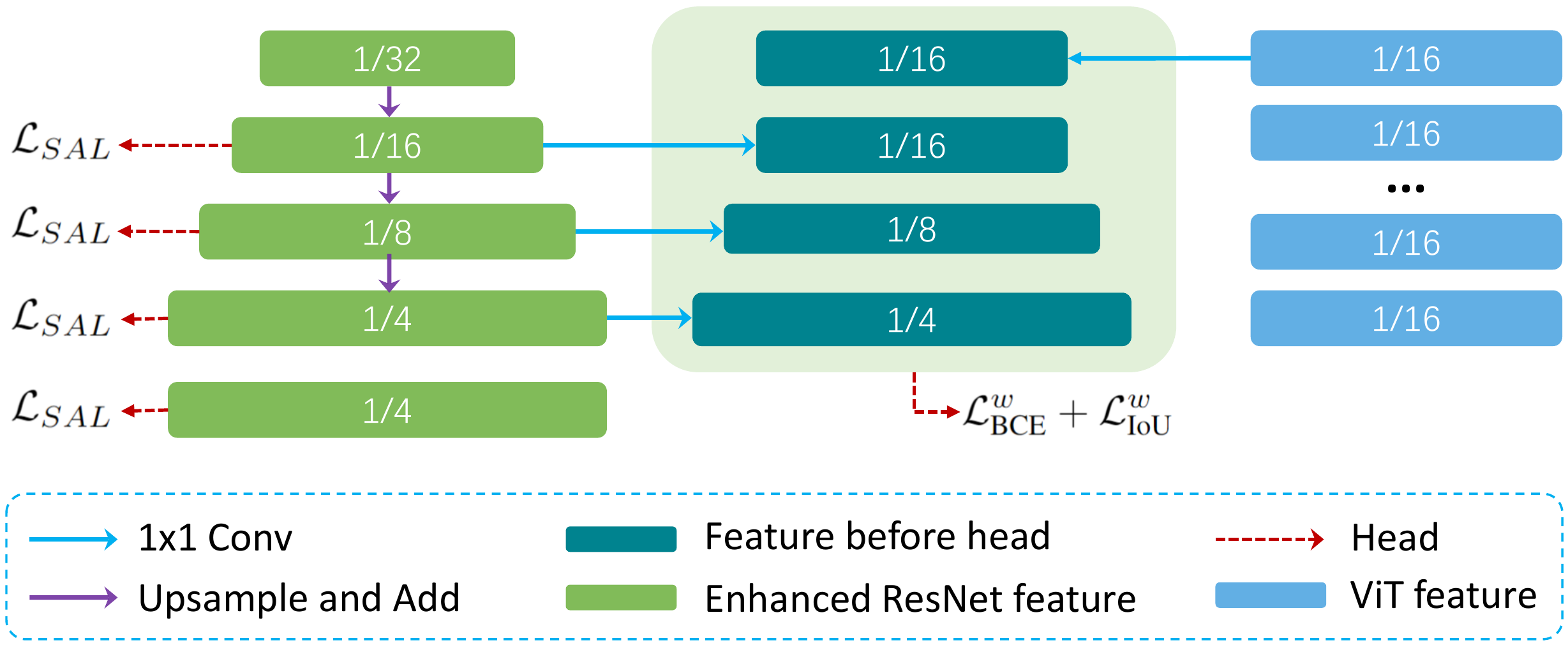}
\caption{Illustration of decoder in DQnet. The enhanced ResNet features after RBQ are combined with ViT feature to form a simple feature pyramid. (Best viewed in color.)}
\label{fig:decoder}
\end{figure}

The decoder networks of our models follow the same framework as shown in \cref{fig:decoder}.

\noindent\textbf{Main Decoder}
For the main experiment, we choose ResNet-50 and ViT-B as feature extractors. For Resnet-50 encoder, the top layer feature is noisy and acounts for large memory overhead, thus we choose four features maps remained after enhancement denoted as $\left\{ \bm{B}_i|i=2,3,4,5 \right\}$, corresponding to scales $\{\frac{1}{4}, \frac{1}{8}, \frac{1}{16}, \frac{1}{32}\}$. Due to the non-hierarchical structure of the vanilla vision transformer, we only adopted the feature maps of the final output, denoted as $\bm{V}_1$, which should have the most representative semantics. Setting the default patch size as 16, the output of ViT is of size $\frac{H}{16}\times\frac{W}{16}$.

For simlicity, the scale fusion is realized by simple top-down and lateral connections. Concretely, 1$\times$1 convolution is used to align the channel dimension to 256, and features of later CNN stages are then upsampled through deconvolution and added to the features of earlier stages progressively, whereas the ViT feature remains identity. Thus we get a simple feature pyramid of scales $\{\frac{1}{4}, \frac{1}{8}, \frac{1}{16}\}$, where the $\frac{1}{16}$ scale covers both $R_4$ and $V_1$. These features are concatenated together and fed into the segmentation head (Upsampling to groundtruth size followed by a 3x3 conv to project the channel dimension from 256 to 1), the whole procedure can be formulated as 
\begin{equation}\footnotesize
\begin{aligned}
{\widetilde{R}} & = {Conv}_{1 \times 1}(R_i), \ \ {\widetilde{V_i}} = {Conv}_{1 \times 1}(V_i)\\
    H_i & = 
    \left\{
    \begin{aligned}
    & {\widetilde{V_i}},  \ \ i=1,\\
    &{\widetilde{R}}_{i} + {Upsample}({\widetilde{R}}_{i+1} ),\ \ i=2,3,4,\\
    \end{aligned}
    \right.\\
F&={Head}\left({Conv}\left(\vert\vert_{i=1}^{4}H_i\right)\right)
\end{aligned}
\end{equation}

\noindent\textbf{Auxiliary Head}
In order to enable the RBQ module to generate more accurate detail querying result $B$ and guide the DQnet to focus on more significant details, an auxiliary head is added to the network for additional output, as shown in \cref{fig:decoder}. To be specific, we let enhanced features $\left\{ \bm{B}_i|i=1,2,3,4 \right\}$ after RBQ getting through a sigmoid layer and a convolution layer, which ends up with four different prediction maps for further supervision with the proposed SAL. 

\subsection{Multi-scale Cross-Model Querying}\label{sec:decoderframework}
The performance of DQnet largely resides in the ability to query details in different scales. The overall algorithm for multi-scale querying is specified in \cref{alg:multi-scale}.
\begin{algorithm}
	\caption{Detail Querying in Multiple Scales}
	\label{alg:multi-scale}
	\begin{algorithmic}[1]
		\Require
		A training images I, a pretrained CNN $\mathcal{C}$ with blocks $\{\mathcal{C}_j\}_{j=1}^{G}$, a pretrained ViT $\mathcal{T}$ and RBQ module $\mathcal{Q}$;
		\Ensure
		$\{B_j\}_{j=1}^{G}$: multi-scale CNN features after enhancement of RBQ;
		$V_1$: ViT final feature;
        \State $V_1 \gets \mathcal{T}(I)$;
		\For{$i \in [1, G]$}
		\If{$i = 1$} \Comment{Block 1}
		\State $R_1 \gets \mathcal{C}_1(I)$;
		\State $B_1 \gets \mathcal{Q}(V_1, R_1)$;
		\Else  \Comment{Block i, $1 < i \leq G$}
		\State $R_i \gets \mathcal{C}_i(B_{i-1})$;
		\State $B_i \gets \mathcal{Q}(V_1, R_i)$;
		\EndIf
		\EndFor
	\end{algorithmic}
\end{algorithm}

{\small
\bibliographystyle{ieee_fullname}
\bibliography{egbib}
}

\end{document}